\documentclass[10pt,twocolumn,letterpaper]{article}

\usepackage[pagenumbers]{cvpr} 

\definecolor{cvprblue}{rgb}{0.21,0.49,0.74}
\usepackage[pagebackref,breaklinks,colorlinks,allcolors=cvprblue]{hyperref}

\def\paperID{*****} 
\def\confName{Arxiv}
\def\confYear{2026}

\title{Mix3R: Mixing Feed-forward Reconstruction and Generative 3D Priors for Joint Multi-view Aligned 3D Reconstruction and Pose Estimation}

\author{Siyou Lin\\
Tsinghua University\\
Beijing, China\\
{\tt\small linsy21@mails.tsinghua.edu.cn}
\and
Zhou Xue\\
Tsinghua University\\
Beijing, China\\
{\tt\small xuezhou08@gmail.com}
\and
Hongwen Zhang\\
Beijing Normal University\\
Beijing, China\\
{\tt\small zhanghongwen@bnu.edu.cn}
\and
Liang An\\
Tsinghua University\\
Beijing, China\\
{\tt\small anliang@mail.tsinghua.edu.cn}
\and
Dongping Li\\
ByteDance\\
Hangzhou, China\\
{\tt\small lidongping83@gmail.com}
\and
Shaohui Jiao\\
ByteDance\\
Beijing, China\\
{\tt\small jiaoshaohui@bytedance.com}
\and
Yebin Liu\\
Tsinghua University\\
Beijing, China\\
{\tt\small liuyebin@mail.tsinghua.edu.cn}
}

\begin{document}
\maketitle

\begin{abstract}
    Recent trends in sparse-view 3D reconstruction have taken two different paths: feed-forward reconstruction (such as VGGT) that predicts pixel-aligned point maps without a complete geometry, and generative 3D reconstruction (such as TRELLIS) that generates complete geometry but often with poor input-alignment. We present \emph{Mix3R}, a novel generative 3D reconstruction method which mixes feed-forward reconstruction and 3D generation into a single framework in an aligned manner. Mix3R generates a 3D shape in two stages: a sparse voxel generation stage and a textured geometry generation stage. Unlike pure generative methods, our first-stage generation jointly produces a coarse 3D structure (sparse voxels), per-view point maps and camera parameters aligned to that 3D structure. This is made possible by introducing a Mixture-of-Transformers architecture that inserts global self-attentions to a feed-forward reconstruction model and a 3D generative model, both pretrained on large-scale data. This design effectively retains the pretrained priors but enables better 2D-3D alignment. Based on the initial aligned generations of sparse 3D voxels and point maps, we compute an overlap-based attention bias that is directly added to another pretrained textured geometry generation model, enabling it to correctly place input textures onto generated shapes in a training-free manner. Our design brings mutual benefits to both feed-forward reconstruction and 3D generation: The feed-forward branch learns to ground its predictions to a generative 3D prior, and conversely, the 3D generation branch is conditioned on geometrically informative features from the feed-forward branch. As a result, our method produces 3D shapes with better input alignment compared with pure 3D generative methods, together with camera pose estimations more accurate than previous feed-forward reconstruction methods. Our project page is at \url{https://jsnln.github.io/mix3r/}
\end{abstract}

\section{Introduction}
\label{sec:intro}

3D reconstruction and camera pose estimation from multi-view images is an important technique for 3D asset acquisition and spatial perception. Traditional multi-view 3D reconstruction methods such as COLMAP~\citep{schoenberger2016sfm,schoenberger2016mvs} require a dense camera set and feature matching techniques to recover the point cloud geometry and camera parameters. Despite their accuracy, these methods usually have high computational complexity and cannot adapt to extreme cases such as sparse views. Recently, two new types of 3D reconstruction have gradually taken over the research trend: feed-forward reconstruction and generative reconstruction.

A common approach taken by feed-forward reconstruction methods (e.g., VGGT~\citep{wang2025vggt}, $\pi^3$~\citep{wang2025pi3}, MapAnything~\citep{keetha2025mapanything} and DepthAnything3~\citep{lin2025da3}) is to directly regress pixel-aligned attributes including depth maps, point maps or ray maps. Camera poses can also be predicted using dedicated decoder heads or inferred from point maps~\citep{lepetit2009epnp}. The pixel-aligned design of feed-forward methods generally produces good input-alignment, but also brings the natural drawback that they only reconstruct the seen part and strongly relies on overlaps between input views. For sparse views, they would produce inaccurate or incomplete reconstructions. On the other hand, 3D generative models~\citep{li2025craftsman3d,yang2024hunyuan3d,hunyuan3d22025tencent,lai2025hunyuan3d25highfidelity3d,xiang2025trellis,xiang2025trellis2,wu2025unilat3d,ye2025hi3dgen,wu2024direct3d,wu2025direct3ds2,li2025sparc3d,chen2025ultra3d} allow modeling multi-view reconstruction as an image-conditioned generation process. Despite the visually pleasant generated shapes, they are merely look-alikes of the input and may not faithfully preserve the geometric dimensions and texture details because they lack explicitly aligned control signals during the injection of image conditions.

A few recent advances attempt to incorporate reconstruction, generation and pose estimation into one single task. ReconViaGen~\citep{chang2025reconviagen} injects VGGT~\citep{wang2025vggt} features into the generation process to improve input alignment. CUPID~\citep{huang2025cupid} jointly generates a UV voxel volume from which camera poses can be solved using PnP. While it demonstrated the ability of a multi-view extension using multi-diffusion~\citep{bar2023multidiffusion}, theoretically this lacks inter-view knowledge to handle rotationally symmetric shapes with asymmetric textures.

Despite these efforts, we would like to ask: How can we unify feed-forward reconstruction and generative reconstruction in a mutually beneficial way? The key difficulty lies in their \emph{mutual alignment}. If a feed-forward model already has the knowledge of the underlying 3D shape, then it can ground its predictions to that shape instead of relying on image overlaps. Conversely, if a generative reconstruction model is conditioned on known camera poses, we can leverage fine-grained pixel alignment to generate 3D shapes that better match the input images. This is a chicken-and-egg problem.

To tackle these issues, we propose \emph{Mix3R}, a novel method that achieves aligned feed-forward reconstruction and generative reconstruction in a mutually beneficial way. Our framework adopts a two-stage coarse-to-fine pipeline following TRELLIS~\citep{xiang2025trellis}. Given multi-view input images of an object, the first stage jointly generates a coarse 3D shape and predicts point maps and camera poses in an aligned manner, and the second stage generates more detailed geometry with input-aligned texture utilizing the alignment from the first stage. In the first stage, we design a mixture-of-transformers (MoT)~\citep{liang2025mot} architecture for TRELLIS~\citep{xiang2025trellis} and $\pi^3$~\citep{wang2025pi3} and train it to predict aligned voxels and point maps. In the second stage, we design an attention bias based on the overlaps between 3D voxels and 2D image patches. In a training-free manner, the attention bias is added on top of a pretrained flow model to generate the final 3D asset but with more accurately aligned control.

Our designs lead to mutual benefits for both feed-forward reconstruction and 3D generation. The information exchange in the MoT architecture allows the feed-forward branch to learn to ground its predictions to a generative 3D prior, and conversely, the 3D generation branch is now conditioned on geometrically informative features from the feed-forward branch. As a result, our method produces 3D assets with better input alignment compared with pure 3D generative methods, together with camera pose estimations more accurate than previous
feed-forward reconstruction methods. In summary, our contributions are as follows.
\begin{itemize}
  \item We propose \emph{Mix3R}, a novel framework that effectively unifies 3D generation and feed-forward reconstruction in an aligned manner, by fusing two pretrained models from each domain to achieve joint geometry generation and camera pose estimation. Unlike ReconViaGen which is a one-way process of injecting VGGT features into generation, our architecture is designed so that they become mutually beneficial.
  \item To best utilize existing pretrained models, we design a mixture-of-transformers (MoT) architecture to incorporate the priors of a feed-forward reconstruction model ($\pi^3$) and a 3D generative model (TRELLIS). Our MoT design mutually benefits both branches in terms of alignment by allowing information exchange between the generative 3D prior and geometrically informative pixel-aligned features.
  \item To further improve alignment in the final textured geometry generation, we propose an attention bias based on the overlaps between the generated coarse voxels and the aligned image patches. The attention bias is added to a pretrained textured geometry flow model in a training-free manner, boosting the quality with minimal extra cost.
\end{itemize}


\section{Related Work}
\label{sec:related-work}

We categorize 3D reconstruction methods into two categories: generative reconstruction which utilizes conditional generative models to produce actual 3D models from images, and feed-forward reconstruction which directly maps one or more input images to 3D.

\subsection{Generative Reconstruction}

There has been a number of 3D generation methods based on different generative frameworks, e.g., GANs~\citep{chan2022eg3d,deng2022gram,gao2022get3d,skorokhodov20233dgp,wu20163dgan,zheng2022sdfstylegan} and diffusion models~\citep{luo2021,schroeppel2024,hu2024,muller2023diffrf,tang2023volumediffusion,chen2023ssdnerf,shue2023,wang2023rodin,zhang2024rodinhd,he2024gvgen,zhang2024gaussiancube}. While some works are purely generative, others allow conditioned generation and can thus be used as a generative reconstruction method. Early methods mainly focus on simple representations such as point clouds~\citep{luo2021,schroeppel2024}, tri-planes~\citep{chan2022eg3d,shue2023,wang2023rodin,zhang2024rodinhd} and volumes~\citep{hu2024,muller2023diffrf,tang2023volumediffusion}, and the quality of their generated models are limited by factors such as point number and volume resolution. As a result, these methods cannot produce asset-level 3D objects.

Recently, inspired by latent diffusion models on images~\citep{rombach2022ldm}, 3D generation also started using latent-space generation. A natural and common choice of latent representation is sparse voxels~\citep{xiang2025trellis,wu2025unilat3d,ye2025hi3dgen,wu2024direct3d,wu2025direct3ds2,li2025sparc3d,chen2025ultra3d}, which, upon generation, can be further decoded to different 3D representations such as meshes, 3D Gaussians~\citep{kerbl20233dgs} or radiance fields~\citep{mildenhall2020nerf}. Another popular choice of compression is VecSet~\citep{zhang2023vecset}, which encodes 3D shapes into vectors for better compression. Building on the VecSet representation, a number of works are capable of generating high-quality 3D assets~\citep{zhang2024clay,li2025craftsman3d,yang2024hunyuan3d,hunyuan3d22025tencent,lai2025hunyuan3d25highfidelity3d}.

Among these works, closest to ours is image-conditioned generative models, e.g., TRELLIS~\citep{xiang2025trellis}. However, methods like TRELLIS only injects images conditions using cross attention modules without explicit view-object alignment. As a result, the generated models may not align well with the input image signal.

\subsection{Feed-forward Reconstruction}

Unlike generative reconstruction which directly models the data distribution (usually embedded with a normalized coordinate system), feed-forward reconstructions directly regresses the target geometry. These methods are mostly \emph{pixel-aligned}. For example, \citet{saito2019pifu,saito2020pifuhd,xiu2022icon,xiu2023econ} use pixel-aligned image features for monocular human reconstruction, achieving good input alignment but often fail for unseen regions. To alleviate this, reconstructions must consider multi-view inputs, which requires either direct or indirect pose estimation. Early work such as FORGE~\citep{jiang2024forge} estimates poses to fuse multi-view features into a unified feature space and then decodes it into a NeRF~\citep{mildenhall2020nerf} volume. Recently, point map regression became the trending paradigm of feed-forward methods. Pioneer work DUSt3R~\citep{wang2024dust3r} directly regresses stereo point maps in a Siamese manner. MonST3R~\citep{zhang2025monst3r} extends it to dynamic scenes using the same paradigm. Subsequent methods CUT3R~\citep{wang2025cut3r} and TTT3R~\citep{chen2026tttr} employs a latent state to further support long-term streaming. Unlike these methods which rely on image pairs or image-latent pairs, VGGT~\citep{wang2025vggt} and its follow-up works~\citep{wang2025pi3,keetha2025mapanything} directly regress point maps and camera poses from an entire image collection. Follow-up works extend this pixel-aligned point map prediction paradigm to predicting 3D Gaussians for rendering purposes~\citep{jiang2025anysplat,xu2024freesplatter,wang2025volsplat}. Despite being able to estimate camera poses from sparse views, these methods only reconstruct seen regions and requires enough image overlap and coverage for good reconstruction quality.

Other than pixel-aligned feed-forward methods, there also exist feed-forward method that do not consider explicit  multi-view alignment. For example, LRM~\cite{hong2024lrm}, LEAP~\citep{jiang2024leap} and PF-LRM~\citep{wang2024pflrm} predict a NeRF~\citep{mildenhall2020nerf} volume from monocular or multi-view images. InstantMesh~\cite{xu2024instantmesh} uses multi-view image diffusion and LRM in a cascaded manner for better multi-view support. SpaRP~\citep{xu2024sparp} utilizes Stable Diffusion~\citep{rombach2022ldm} to generate multi-view point maps and RGB images at given poses with unconstrained input views, turning pose-free reconstruction into traditional MVS with known cameras. Due to the lack of explicit input-output alignment and stochastic modeling of unseen regions, these methods often produces less aligned geometries or blurry textures.

\subsection{Unifying Reconstruction and Generation}

Attempts have been made to unify feed-forward reconstruction and generation. CAST~\citep{yao2024cast} and SAM3D~\citep{sam3dteam2025sam3dobjects} can be conditioned on depth maps to generate shapes and estimate poses that closely match the input images. However, both CAST and SAM3D focus on single-view multi-object composition, not multi-view alignment and fusion. Recently, ReconViaGen~\cite{chang2025reconviagen} incorporates multi-view reconstruction guidance from the feed-forward model VGGT~\cite{wang2025vggt} to further enhance alignment. Concurrently, CUPID~\cite{huang2025cupid} jointly generates a UV voxel grid to solve for view-object alignment and use it for better texture generation. Focusing on scenes, Gen3R~\citep{huang2026gen3r} and Aether~\citep{zhu2025aether} also explored unifying video generation and dynamic reconstruction in the sense that the former can be repurposed to achieve the latter. However, these are less related to our object-centric setting.

\section{Method}

\begin{figure*}[t!]
  \centering
   \includegraphics[width=\linewidth]{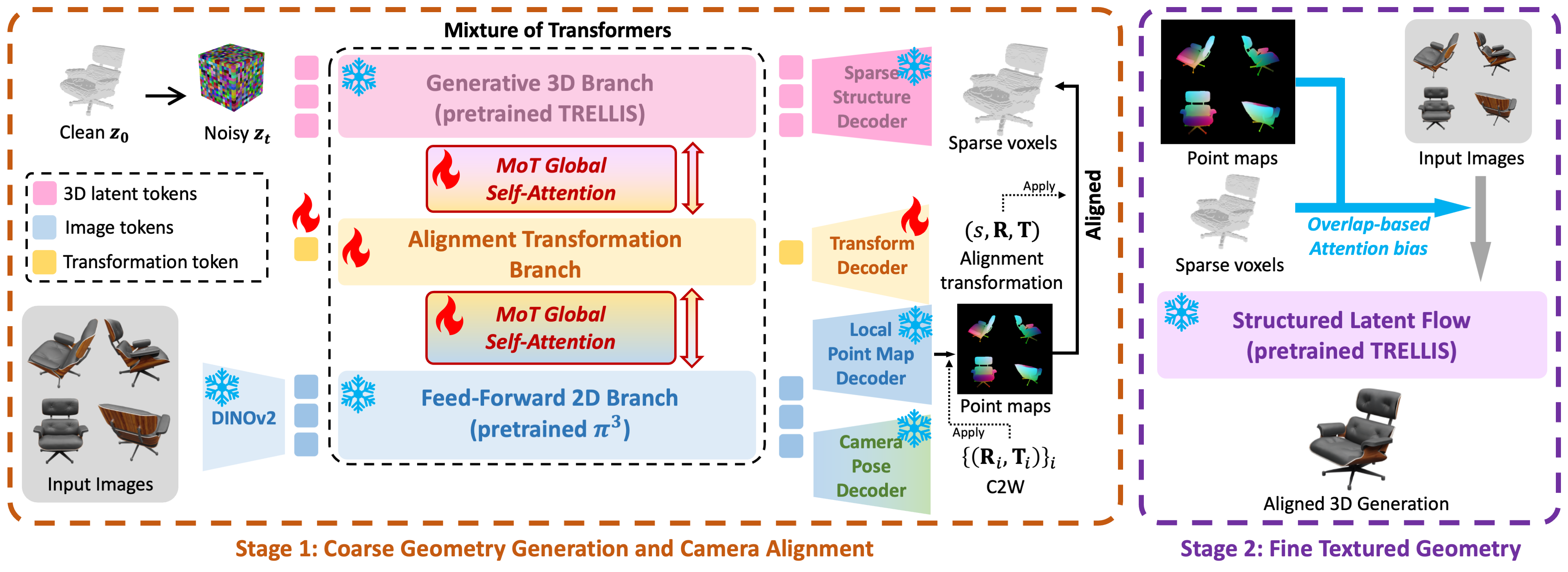}
   \caption{The overall architecture of our two-stage framework. Given multi-view unposed input images, we first employ a mixture-of-transformers architecture that jointly infers a coarse 3D structure, pixel-aligned local point maps, camera poses, and an alignment transformation that aligns point maps to the 3D shape. This alignment is then used to provide fine-grained control in the form of attention bias for the final 3D asset generation.}
   \label{fig:architecture}
\end{figure*}

\subsection{Preliminaries: TRELLIS and $\pi^3$}
\label{sec:preliminaries}

Our goal is to achieve 3D generation and camera pose estimation from multi-view images in a unified and aligned manner. We build our model upon two existing models with large-scale pretraining: TRELLIS~\citep{xiang2025trellis} and $\pi^3$~\citep{wang2025pi3}. We begin with a short introduction to them.

TRELLIS is a 3D generative model based on flow matching~\citep{lipman2023flow}. TRELLIS models each 3D shape using a structured latent representation: a set of features $\{\mathbf f_i\}_{i=1}^L$ attached to the non-zero voxels $\{\mathbf p_i\}_{i=1}^L$ of an occupancy grid $\mathbf O\in\mathbb\{0,1\}^{64^3}$. For simplicity, we use $\mathbf f=\{(\mathbf f_i, \mathbf p_i)\}$ to represent a structured latent. Each $\mathbf f$ can be further decoded into a mesh surface, a 3DGS~\citep{kerbl20233dgs} point cloud or a NeRF~\citep{mildenhall2020nerf} using dedicated decoders. TRELLIS generates a 3D asset in two stages. First, a sparse structure latent code ${\mathbf z}\in{\mathbb R}^{16^3\times8}$ is generated using a flow transformer $\mathcal F_{\rm ss}$. Then, a sparse structure decoder $\mathcal D_{\rm ss}$ decodes ${\mathbf z}$ into an occupancy grid $\mathbf O=\mathcal D_{\rm ss}(\mathbf z)$. Non-empty voxels $\{\mathbf p_i\}_{i=1}^L$ are then extracted from $\mathbf O$. A second sparse flow transformer $\mathcal F_{\rm slat}$ then generates a structured latent code $\mathbf f$ on $\{\mathbf p_i\}_{i=1}^L$. Both $\mathcal F_{\rm ss}$ and $\mathcal F_{\rm slat}$ use cross-attention to inject image conditions and employ standard Euler sampler during generation.

$\pi^3$~\citep{wang2025pi3} is a feed-forward reconstruction method. Given images $\{\mathbf I_i\}_{i=1}^N$, $\pi^3$ processes them using a permutation invariant vision transformer and obtain camera-space point maps $\mathbf X_i\in\mathbb R^{H\times W\times 3}$ camera poses in the form of camera-to-world transformations $(\mathbf R_i, \mathbf T_i)\in{\rm SE}(3)$.
World-space point maps can be obtained as $\mathbf R_i(\mathbf X_i)+\mathbf T_i$, where $(\mathbf R_i, \mathbf T_i)$ applies to each pixel in $\mathbf X_i$ independently.
Since $\pi^3$ has an input permutation invariant design, and $\mathbf X_i, \mathbf R_i, \mathbf T_i$ are trained with affine-invariant losses, the output distribution of $\pi^3$ is generally more stable compared with reference frame-based methods such as VGGT~\citep{wang2025vggt}.

\subsection{Joint Coarse Geometry Generation and Camera Pose Estimation}

The overall architecture is shown in Fig.~\ref{fig:architecture}. Our stage-1 model jointly generates a coarse 3D structure and camera poses aligned to it. To best utilize existing pretrained models, we choose the TRELLIS sparse structure flow model as our generative branch (3D branch) and the $\pi^3$ backbone transformer as our feed-forward branch (2D branch). Note that TRELLIS and $\pi^3$ have different output coordinate spaces. Thus, we add an extra transformation branch and a decoder head to predict a similarity transform $(s,\mathbf R,\mathbf T)$ that aligns the output of $\pi^3$ to the voxel space of TRELLIS. These branches are incorporated into a single large transformer using the MoT paradigm~\citep{liang2025mot}. We will explain the inputs and outputs of the model in this section. Specific architectural designs are deferred to Sec~\ref{sec:method:mot-arch-design}.

Let $\mathcal F_{\rm mix}$ denote our stage-1 network. For notational simplicity, we assume the DINOv2 encoder~\citep{oquab2023dinov2,darcet2023vitneedreg,jose2024dinov2meetstextunified}, the main transformer blocks and all the decoders shown in Fig.~\ref{fig:architecture} are represented by $\mathcal F_{\rm mix}$. The network $\mathcal F_{\rm mix}$ takes the following inputs: (1) a noisy sparse structure latent code
\begin{equation}
    \mathbf z_t=(1-t)\mathbf z_0+t\mathbf\epsilon, t\in[0,1]
\end{equation}
where $\mathbf z_0$ is its corresponding clean latent code, $\epsilon$ is a Gaussian noise vector having the same dimensions, and $t$ is the time step used in standard flow matching~\citep{lipman2023flow}; (2) multi-view images $\{\mathbf I_i\}_{i=1}^N$ of the 3D shape corresponding to $\mathbf z_0$; (3) A learnable token $\mathbf g$ representing the alignment transformation $(s,\mathbf R,\mathbf T)$. Following the flow matching paradigm~\citep{lipman2023flow}, the 3D branch predicts a velocity $\mathbf v$, which is trained to match $\mathbf\epsilon-\mathbf z_0$. The image tokens of $\{\mathbf I_i\}_{i=1}^N$ are processed by the 2D branch and then decoded into local point maps $\{\mathbf X_i\}_{i=1}^N$ and camera poses $\{(\mathbf R_i,\mathbf T_i)\}_{i=1}^N$. We can write down the whole network as:
\begin{equation}
    \mathbf v, \{\mathbf X_i\}, \{(\mathbf R_i,\mathbf T_i)\}, (s,\mathbf R,\mathbf T)=\mathcal F_{\rm mix}(\mathbf z_t,t,\{\mathbf I_i\},\mathbf g).
\end{equation}

For training, $\mathbf v$ is supervised by the standard flow matching loss
\begin{equation}
    \mathcal L_{\rm fm}=\|\mathbf v-(\mathbf\epsilon-\mathbf z_0)\|^2.
\end{equation}
For the output point maps, camera poses and the alignment transformation, instead of supervising them separately, we first compute the point maps after applying these transformations:
\begin{equation}
    \hat{\mathbf X}_i=s(\mathbf R(\mathbf R_i(\mathbf X_i)+\mathbf T_i)+\mathbf T).\label{eq:aligned-pointmaps}
\end{equation}
All the aligned point maps $\{\hat{\mathbf X}_i\}$ are supervised by two losses $\mathcal L_{\rm pts}$ and $\mathcal L_{\rm nml}$ between them and the ground truth point maps. Here, $\mathcal L_{\rm pts}$ is the L1 loss on point coordinates, and $\mathcal L_{\rm nml}$ is the L1 loss on point normals computed from point maps. The final loss is
\begin{equation}
    \mathcal L=\mathcal L_{\rm fm}+\lambda_{\rm pts}\mathcal L_{\rm pts}+\lambda_{\rm nml}\mathcal L_{\rm nml}.
\end{equation}

\begin{figure*}[t!]
  \centering
   \includegraphics[width=\linewidth]{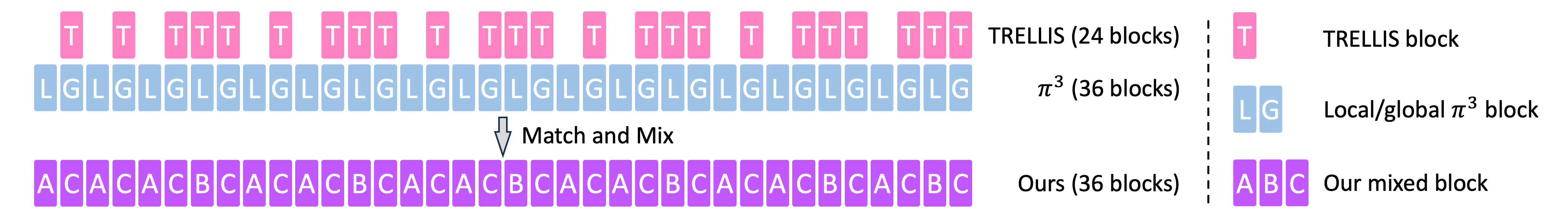}
   \caption{The block matching configuration of our MoT architecture. According to different matching types, our network has three different types of mixed blocks.}
   \label{fig:block-matching}
\end{figure*}

\begin{figure}[t!]
  \centering
   \includegraphics[width=\linewidth]{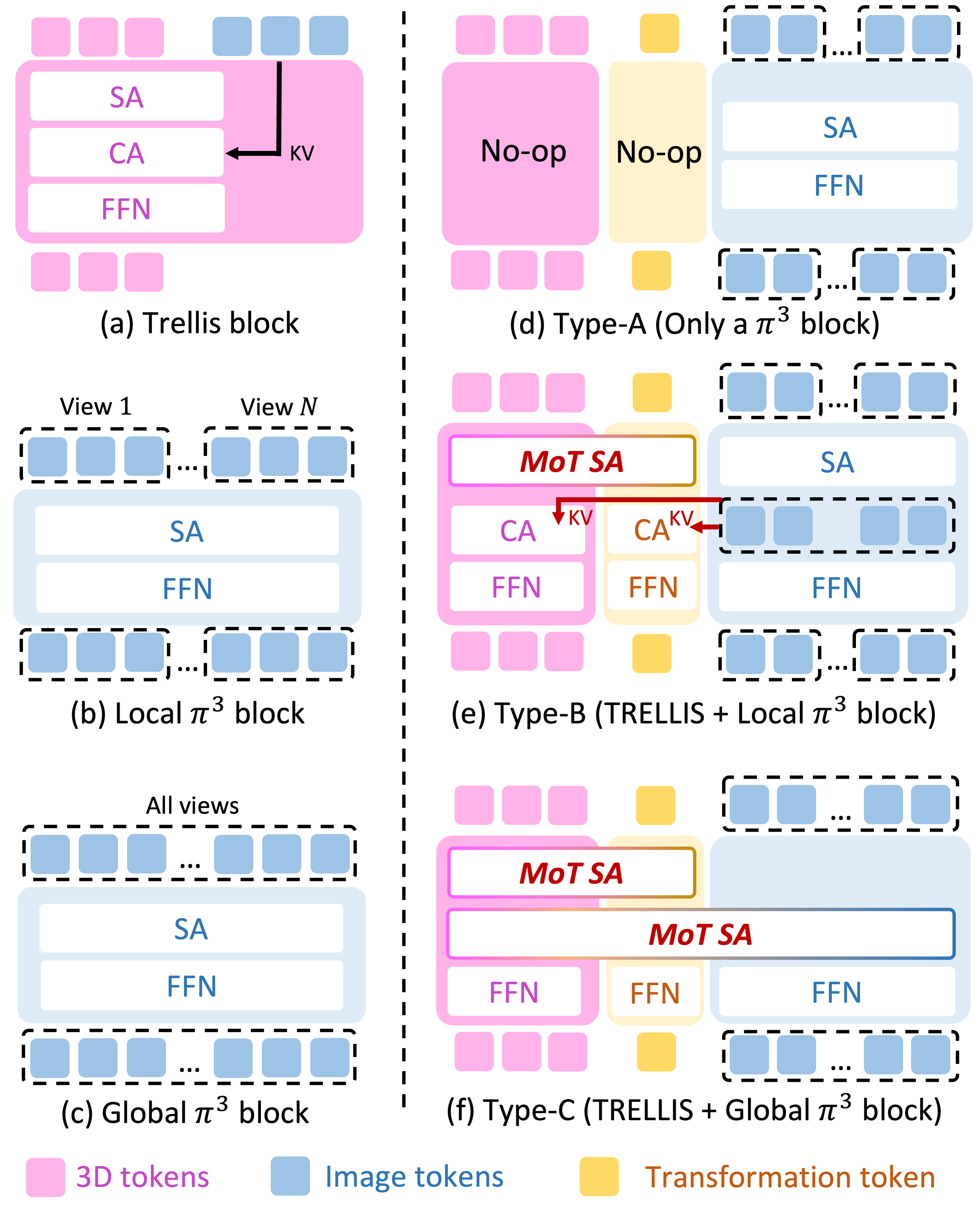}
   \caption{Illustrations of different block mixture architectures. Sub-figures (a), (b) and (c) on the left show the structures of the original TRELLIS blocks and $\pi^3$ blocks, whereas (e), (f) and (g) show the three types of mixed blocks obtained from our block matching strategy in Sec.~\ref{sec:method:mot-arch-design}. Note that we still use residual connections, layer normalization, time step modulation and QK-norm, but do not show them in this figure for simplicity.}
   \label{fig:architecture-blocks}
\end{figure}

Note that our goal is to align point maps $\{\hat{\mathbf X}_i\}$ to the shape $\mathbf z_0$. However, since the network $\mathcal F_{\rm mix}$ only sees the noisy version $\mathbf z_t$. When $t$ is large, almost no geometric information is retained in $\mathbf z_t$ and the losses $\mathcal L_{\rm pts}$, $\mathcal L_{\rm nml}$ become ambiguous. Therefore, we choose their coefficients to depend on $t$, and empirically set them as
\begin{equation}
    \lambda_{\rm pts}={\rm Sigmoid}(-24t+9),\ \lambda_{\rm nml}=0.1\times\lambda_{\rm pts}.
\end{equation}
Note that this implies $t\geq0.5\implies\lambda_{\rm pts}\approx0$ and $t\leq0.25\implies\lambda_{\rm pts}\approx1$. This choice is based on an empirical observation that when $t\geq0.5$ almost no geometry can be recovered from $\mathbf z_0$ using the sparse structure decoder of TRELLIS, while when $t\leq0.25$ the decoded geometry is mostly complete except for minor details.

\subsection{Architectural Designs of the Mixture-of-Transformers Network}
\label{sec:method:mot-arch-design}

In this section, we explain our specific architectural designs. Our MoT network is a mixture of two pretrained models: the sparse structure flow transformer in TRELLIS~\citep{xiang2025trellis} and the backbone transformer of $\pi^3$~\citep{wang2025pi3}. Fig.~\ref{fig:architecture-blocks}(a,b,c) show the original block structures of TRELLIS and $\pi^3$, respectively. Note that $\pi^3$ follows VGGT~\citep{wang2025vggt} to alternate between local self-attention (different views are batched) and global self-attention (all views concatenated into a single token sequence). Thus, $\pi^3$ has two types of blocks, local and global, as shown in Fig.~\ref{fig:architecture-blocks}(b,c).

We intend to design a mixed architecture that allows information exchange between TRELLIS and $\pi^3$ by inserting self-attentions between them. While there might be different ways to fuse these two networks, enumerating them is not practical. Instead, we adopt a mixing scheme which best preserves pretrained weights, based on the following principles. (1) To retain the pretrained abilities of TRELLIS and $\pi^3$ as much as possible (trying not to discard pretrained weights), we insert MoT~\citep{liang2025mot} self-attentions since it uses different query/key/value matrices for different modalities. (2) To keep the alternating local/global attention design of $\pi^3$, we only extend the self-attention in $\pi^3$'s global blocks to MoT self-attentions, while local blocks remain local. (3) For simplicity, blocks for the alignment transformation branch adopts a symmetric structure to TRELLIS blocks.

Since TRELLIS has 24 blocks while $\pi^3$ has 36 blocks, we need to define a block matching before inserting MoT attentions between them. Based on principle (2) above, we first guarantee every global $\pi^3$ block is matched with a TRELLIS block by computing a uniform injection from the 18 global $\pi^3$ blocks into the 24 TRELLIS blocks. Then, 6 TRELLIS blocks remain unmatched, but there is only a unique way to match them with the remaining local $\pi^3$ blocks in an order-preserving way (see the supplementary material for details). Finally, this matching scheme leads to the exact matching shown in Fig.~\ref{fig:block-matching}, with 3 types of block mixtures. Fig.~\ref{fig:architecture-blocks}(d) shows type-A matching, where a local $\pi^3$ block is not matched with any TRELLIS block, in which case no mixing actually happens. Fig.~\ref{fig:architecture-blocks}(e) shows type-B matching, where a local $\pi^3$ block is matched with a TRELLIS block. In this case, we extend the original TRELLIS block self-attention to an MoT self-attention across 3D tokens and the transformation token. We also inject the intermediate geometry-informative token features of $\pi^3$ into the 3D branch and the transformation branch using cross-attention modules. Fig.~\ref{fig:architecture-blocks}(f) shows type-C matching, where a global $\pi^3$ block is matched with a TRELLIS block. In this case we further extend the original cross-attention module to a large global MoT self-attention which processes all three modalities at once.

During training, we freeze all parameters that have a pretrained value and are not affected by our newly added modules. Other parameters are activated (see supplementary material for details). In this way, we retain the abilities of the pretrained models as much as possible but also enable different modalities to interact with each other to achieve a good alignment between the feed-forward reconstruction branch and the generation branch.

\subsection{Attention Bias for Training-Free Tuning of Textured Geometry Generation and Camera Refinement}

Given multi-view images $\{I_i\}_{i=1}^N$, we can use our model to jointly generate a sparse structure latent $\mathbf z$, camera poses $\{(\mathbf R_i,\mathbf T_i)\}_{i=1}^N$ and the alignment transformation $(s, \mathbf R,\mathbf T)$. Our next step is to utilize the 2D-3D alignment to generate view-aligned fine geometry and texture.

For the latent code $\mathbf z$, we first use pretrained TRELLIS sparse structure decoder $\mathcal D_{\rm ss}$ to obtain the corresponding occupancy grid $\mathbf O=\mathcal D_{\rm ss}(\mathbf z)$ and extract its non-zero voxels $\{\mathbf p_i\}_{i=1}^L$. Then, following TRELLIS, a second flow transformer $\mathcal F_{\rm slat}$ takes a noisy structured latent as input and gradually denoises it to generate the final clean latent as described in Sec.~\ref{sec:preliminaries}. The image conditions $\{\mathbf I_i\}_{i=1}^N$ are first encoded using DINOv2~\citep{oquab2023dinov2,darcet2023vitneedreg,jose2024dinov2meetstextunified} and then injected into $\mathcal F_{\rm slat}$ using cross-attention modules. Let us denote by $\mathbf f$ the intermediate token set corresponding to the structured latent to be denoised, and denote by $\mathbf y$ the token set of the input DINOv2 tokens. Then a pretrained cross-attention can be written as follows:
\begin{eqnarray}
    &&{\rm CrossAttn}(Q(\mathbf f), K(\mathbf y), V(\mathbf y))\nonumber\\
    &=&{\rm Softmax}\left(\frac{Q(\mathbf f)K(\mathbf y)^T}{\sqrt{d}}\right)V(\mathbf y),
\end{eqnarray}
where $d$ is the feature dimension. We attempt to find a bias matrix $\mathbf B(\mathbf f,\mathbf y)$ such that the modified attention
\begin{equation}
    {\rm Softmax}\left(\frac{Q(\mathbf f)K(\mathbf y)^T}{\sqrt{d}}+\mathbf B(\mathbf f, \mathbf y)\right)V(\mathbf y)
\end{equation}
makes tokens in $\mathbf f$ attend more to relevant tokens in $\mathbf y$.

\begin{figure*}[t!]
  \centering
   \includegraphics[width=\linewidth]{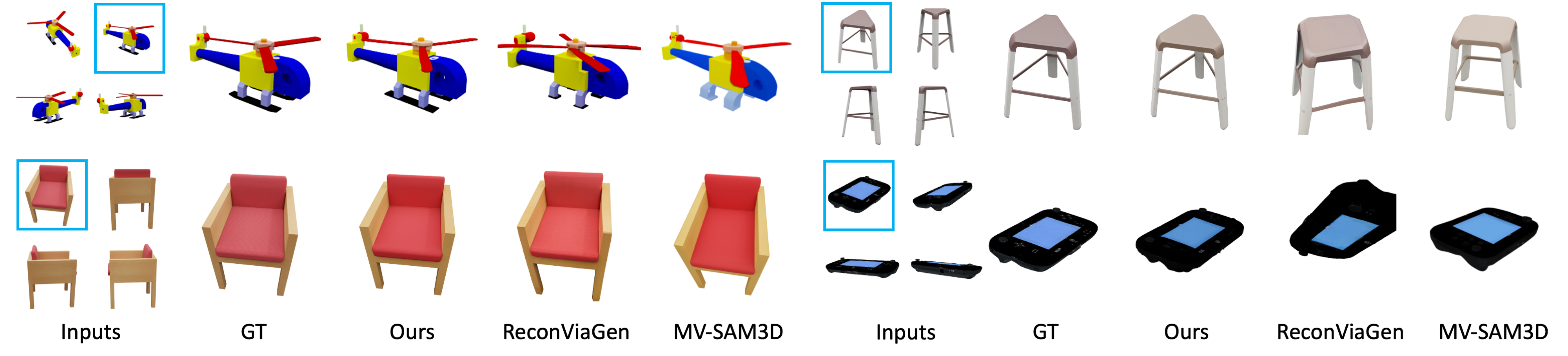}
   \caption{We exhibit the reprojection alignment. Each rendering result is obtained using the decoded 3D Gaussians and the predicted camera parameters.}
   \label{fig:alignment-eval}
\end{figure*}

Note that each token $\mathbf f_j$ in $\mathbf f$ corresponds to a voxel $\mathbf p_j$ while each token $\mathbf y_k$ in $\mathbf y$ corresponds to an image patch of one of $\{I_i\}$. Let $\{\hat{\mathbf X}_i\}$ be the aligned point maps as described in Eq.~\eqref{eq:aligned-pointmaps}, and let $\hat{\mathbf x}_k$ be the point set corresponding to the patch $\mathbf y_k$ in $\{\hat{\mathbf X}_i\}$. For each voxel $\mathbf p_j$, we define its average point count (APC) as
\begin{eqnarray}
    &&{\rm APC}(\mathbf p_j)\nonumber\\
    &=&\left\{
    \begin{array}{ll}
    \frac{\sum_k|\mathbf p_j\cap\hat{\mathbf x}_k|}{\#\{k:|\mathbf p_j\cap\hat{\mathbf x}_k|>0\}},&|\mathbf p_j\cap\hat{\mathbf x}_k|>0\textrm{\ for some\ } k\\
    0,&\textrm{otherwise}
    \end{array}\right.
\end{eqnarray}
where $|\mathbf p_j\cap\hat{\mathbf x}_k|$ is the number of points in $\hat{\mathbf x}_k$ that are contained in voxel $\mathbf p_j$. The final attention bias $\mathbf B(\mathbf f_j,\mathbf y_k)$ added to the score between $\mathbf f_j$ and $\mathbf y_k$ is
\begin{equation}
    \mathbf B(\mathbf f_j,\mathbf y_k)=\alpha\max\left(\frac{|\mathbf p_j\cap\hat{\mathbf x}_k|-{\rm APC}(\mathbf p_j)}{\max_k(|\mathbf p_j\cap\hat{\mathbf x}_k|)-{\rm APC}(\mathbf p_j)},0\right).\label{eq:attn-bias}
\end{equation}
where $\alpha>0$ is a scaling hyperparameter. In our experiments we choose $\alpha=5$.

The idea behind Eq.~\eqref{eq:attn-bias} is that the attention score for $(\mathbf f_j,\mathbf y_k)$ should be increased if $\mathbf y_k$ has an overlap with $\mathbf f_j$ above average among all image tokens. We also empirically found that decreasing the attention scores leads to degraded performance, and thus clip the biases to have a minimum of $0$. Finally, these modified attention scores are used to tune the behavior of $\mathcal F_{\rm slat}$ in a training-free manner to generate the final structured latent, which is then decoded into a mesh and a 3DGS representation.

Recall that our model in stage 1 only predicts camera poses, but not the intrinsics. To further refine the camera parameters, we use the predicted poses as initial values and estimate the intrinsics by solving the perspective projection equation in least squares. Then, the intrinsics and extrinsics are jointly refined using the DRTK differentiable renderer~\citep{pidhorskyi2024drtk} by minimizing the RGB loss and the mask loss between the rendered 3D mesh and input images. More details are in the supplementary material.

\section{Experiments}

\subsection{Implementation Details}

Our model is trained on a subset of the TRELLIS-500k dataset~\citep{xiang2025trellis}, containing 404354 objects. Due to the extreme high cost of the original TRELLIS data processing pipeline, we use TRELLIS to generate our training dataset. For training, we adopt pretrained weights whenever possible and only activate the parameters without a pretrained weight and those related to our newly added MoT self-attentions and cross-attentions. We use a learning rate of $10^{-4}$ with cosine scheduling and train for 400k steps with a batch size of 16. Please refer to the supplementary material for more details.

\begin{figure*}[t!]
  \centering
   \includegraphics[width=\linewidth]{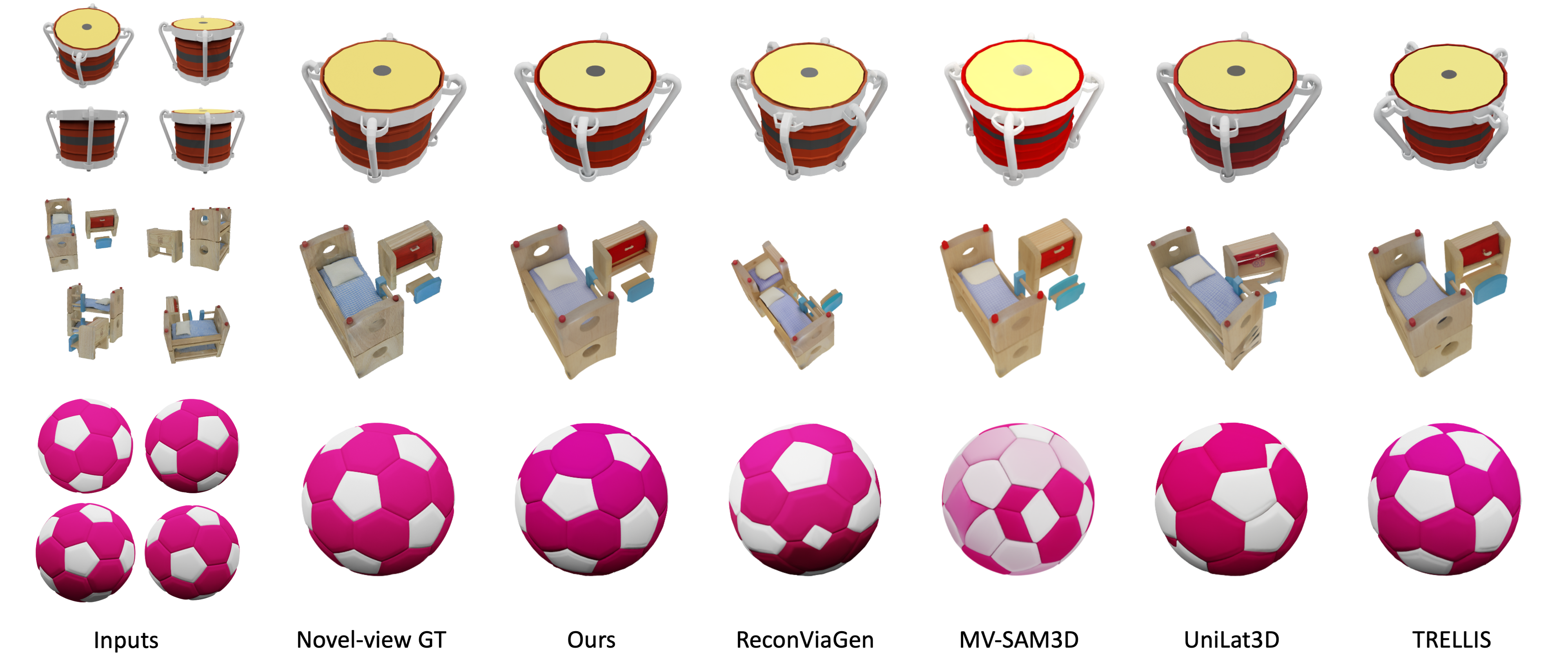}
   \caption{Qualitative results of novel-view rendering evaluation. We show input images and novel-view GT images. Our method more accurately restores texture and geometry.}
   \label{fig:nvs-eval}
\end{figure*}


\begin{table}[h!]
\caption{Metric evaluations for view-object alignment on the Toys4K and the GSO dataset.\label{table:alignment-eval}}
\centering
\small
\begin{tabular}{c|ccc}
\toprule
Dataset     & \multicolumn{3}{c}{Toys4k} \\
\midrule
Metric      & PSNR & SSIM & LPIPS\\
\midrule
MV-SAM3D    & 22.360 & 0.9168 & 0.065 \\
ReconViaGen & 23.834 & 0.9359 & 0.047 \\
Ours        & \textbf{28.144} & \textbf{0.9621} & \textbf{0.030} \\
\midrule
Dataset     & \multicolumn{3}{c}{GSO} \\
\midrule
Metric      & PSNR & SSIM & LPIPS \\
\midrule
MV-SAM3D    & 22.345 & 0.9016 & 0.084 \\
ReconViaGen & 24.036 & 0.9260 & 0.058 \\
Ours        & \textbf{25.031} & \textbf{0.9281} & \textbf{0.057} \\
\bottomrule
\end{tabular}
\end{table}

\begin{table*}[h]
\caption{Novel view synthesis and geometry evaluation.\label{table:nvs-cd-eval}}
\centering
\begin{tabular}{c|cccc|cccc}
\toprule
Dataset     & \multicolumn{4}{c|}{Toys4k} & \multicolumn{4}{c}{GSO} \\
\midrule
Metric      & PSNR & SSIM & LPIPS & CD ($\times10^{-3}$) & PSNR & SSIM & LPIPS & CD ($\times10^{-3}$) \\
\midrule
TRELLIS     & 23.930 & 0.9311 & 0.055 & 7.9986 & 22.420 & 0.9034 & 0.079 & 4.5658 \\
UniLat3D    & 25.082 & 0.9415 & 0.049 & 7.0876 & 22.765 & 0.9102 & 0.076 & 7.1384 \\
MV-SAM3D    & 23.550 & 0.9273 & 0.059 & 2.7556 & 22.331 & 0.9032 & 0.085 & 5.2356 \\
ReconViaGen & 24.491 & 0.9335 & 0.046 & 1.9419 & 24.183 & 0.9152 & 0.060 & 1.0015 \\
Ours        & \textbf{27.177} & \textbf{0.9551} & \textbf{0.033} & \textbf{0.7419} & \textbf{24.826} & \textbf{0.9271} & \textbf{0.055} & \textbf{0.7945} \\
\bottomrule
\end{tabular}
\end{table*}


\subsection{Evaluation}
\label{sec:eval}

Our work aims at jointly generating 3D objects and their alignment to input images. We evaluate our method over three aspects: (1) input alignment; (2) geometry and texture accuracy; (3) camera pose accuracy; (4) real-world phone captures. We use Toys4k~\citep{stojanov21cvpr} and Google Scanned Objects (GSO)~\citep{downs2022gso} as evaluation datasets.

\paragraph{Input alignment}

In this experiment, we directly render the generated 3D models using predicted camera poses to measure how well the generated shape aligns with the input. This experiment simultaneously evaluates the quality of the generated shape and the accuracy of predicted camera poses, since both need to be accurate for the rendered images to be aligned with input images. We compare with ReconViaGen~\citep{chang2025reconviagen} and MV-SAM3D~\citep{li2026mvsam3d} since they both simultaneously generate geometry and pose, and both support multi-view inputs. Please refer to the supplementary material for baseline settings. Fig.~\ref{fig:alignment-eval} shows the qualitative results of the alignment evaluation. Our method not only generates geometry which is aligned with inputs but also estimates correct camera poses which allows accurate reprojection of the 3D shape back to input images, whereas the baseline methods ReconViaGen and MV-SAM3D can generate incorrect structures or wrong poses. We also report the metrics PSNR, SSIM and LPIPS~\citep{zhang2018lpips} in Table~\ref{table:alignment-eval}. Our method performs the best in terms of input-alignment on both benchmark datasets.

\paragraph{Geometry and texture accuracy}

\begin{figure*}
  \centering
   \includegraphics[width=\linewidth]{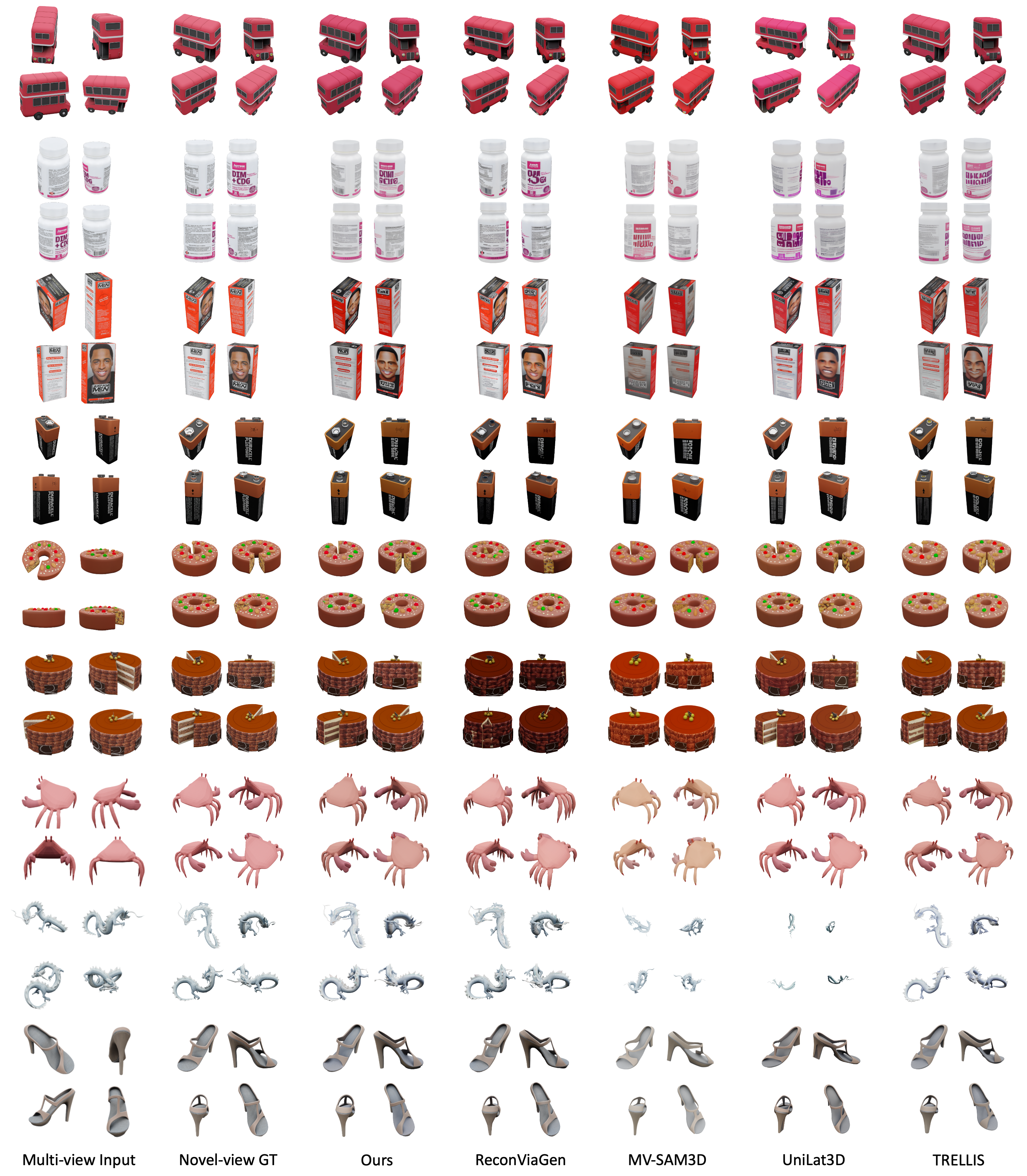}
   \caption{More qualitative results of novel-view rendering evaluation.}
   \label{fig:nvs-eval-v5}
\end{figure*}

\begin{figure*}
  \centering
   \includegraphics[width=0.95\linewidth]{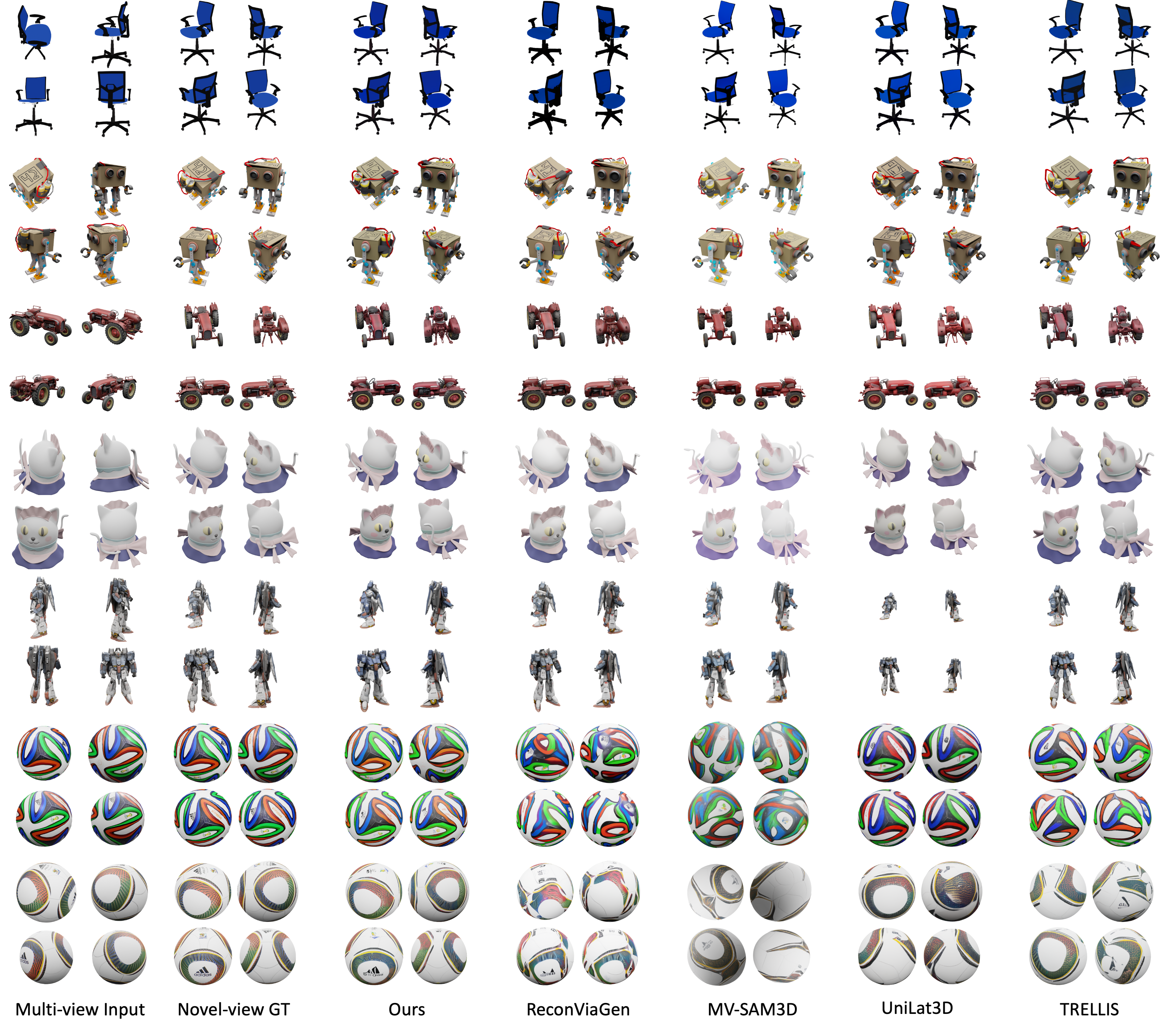}
   \caption{More qualitative results of novel-view rendering evaluation.}
   \label{fig:nvs-eval-v6}
\end{figure*}

We also evaluate the quality of the generated 3D assets, regardless of the predicted camera poses. In this experiment, we additionally compare with TRELLIS~\citep{xiang2025trellis}, UniLat3D~\citep{wu2025unilat3d}. All metrics are computed after a similarity alignment to GT (see the supplementary material for details). For evaluating geometric accuracy, we measure the Chamfer distance (CD). For texture evaluation, we choose 4 novel views different from input views, and measure PSNR, SSIM and LPIPS for all novel-view renderings. Table~\ref{table:nvs-cd-eval} reports the quantitative results. Our generations score the best in terms of both geometry and texture accuracy. Fig.~\ref{fig:nvs-eval} shows some qualitative examples. Our lower Chamfer distance results indicate better preservation of object dimensions and their proportions, even though the visual appearances are sometimes similar to baseline methods. Our method also correctly places an asymmetric input texture onto symmetric shapes, whereas other methods are more likely to generate a plausible but misaligned texture. More results are presented in Fig.~\ref{fig:nvs-eval-v5} and Fig.~\ref{fig:nvs-eval-v6}.

\paragraph{Camera pose accuracy} We compare the accuracy of our camera pose estimation with ReconViaGen, MV-SAM3D, VGGT and $\pi^3$. Since camera poses can be ambiguous up to a similarity transformation, we evaluate the relative rotation accuracy (RRA), relative translation accuracy (RTA) and the area under curve (AUC) with an angle threshold of 30 degrees following \citet{wang2023pd}. Table~\ref{table:camacc-eval} shows the evaluation results, where our method has the best overall performance compared with both feed-forward methods and generative methods.


\begin{table}
\caption{Quantitative evaluations of camera accuracy in terms of relative rotation accuracy (RRA), relative translation accuracy (RTA) and area under curve (AUC). All metrics use a threshold of 30 degrees.\label{table:camacc-eval}}
\centering
\small
\begin{tabular}{c|ccc}
\toprule
Dataset     & \multicolumn{3}{c}{Toys4k} \\
\midrule
Acc@30      & RRA & RTA & AUC \\
\midrule
VGGT        & 93.05 & 88.58 & 57.13 \\
$\pi^3$     & 91.71 & 88.72 & 57.72 \\
MV-SAM3D    & 65.73 & 67.22 & 31.88 \\
ReconViaGen & 82.63 & 85.04 & 49.83 \\
Ours        & \textbf{95.49} & \textbf{95.41} & \textbf{70.63} \\
\midrule
Dataset     & \multicolumn{3}{c}{GSO} \\
\midrule
Acc@30      & RRA & RTA & AUC \\
\midrule
VGGT        & \textbf{97.78} & 91.36 & 60.20 \\
$\pi^3$     & 96.42 & 91.93 & 61.41 \\
MV-SAM3D    & 53.45 & 56.18 & 24.35 \\
ReconViaGen & 91.78 & 91.99 & 57.29 \\
Ours        & 93.50 & \textbf{94.50} & \textbf{66.93} \\
\bottomrule
\end{tabular}
\end{table}

\begin{figure*}
  \centering
   \includegraphics[width=0.95\linewidth]{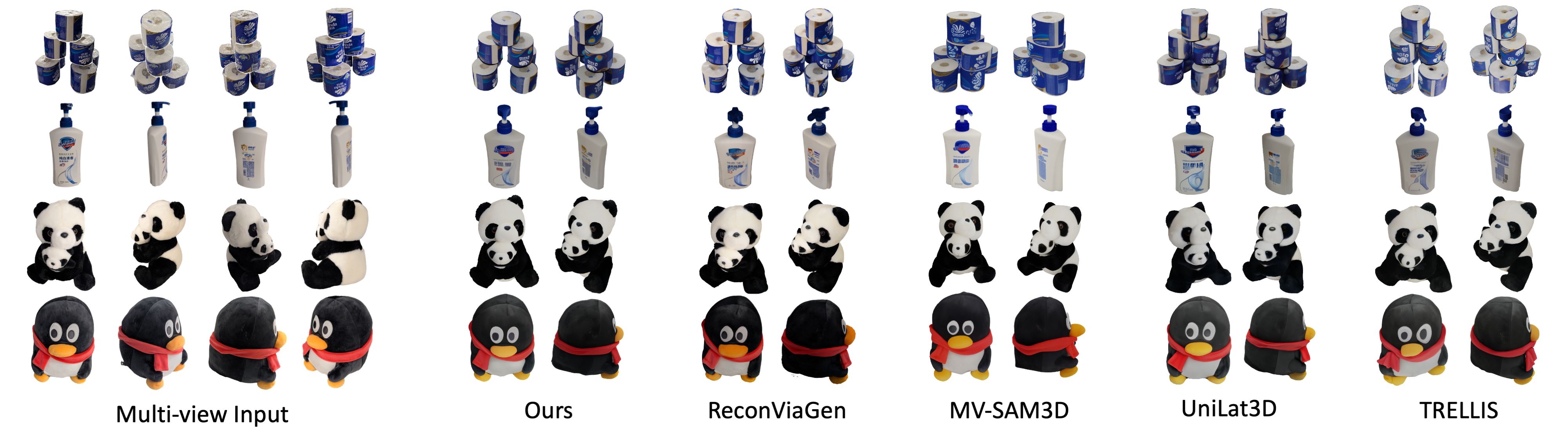}
   \caption{Qualitative results for real-world cellphone captures.}
   \label{fig:realcap}
\end{figure*}

\paragraph{Real-world phone capture} Fig.~\ref{fig:realcap} shows the generation results for real-world objects captured by cellphones. Our method performs comparably with ReconViaGen, while other methods generally suffer from either geometry or texture distortions.

\section{Summary}

In this work, we propose Mix3R, a mixture of a pretrained 3D generative model and a 2D pixel-aligned feed-forward reconstruction model based on the MoT architecture~\citep{liang2025mot}. Our model can jointly generate a sparse voxel structure and point maps aligned to it. Overlaps between the voxel structure and input images can be computed using the aligned point maps as an intermediary. Based on the availability of the overlap, we further compute an attention bias matrix such that the final geometry and texture generation attention correctly to different regions of input images. In this way, we successfully improve the input-alignment of generated 3D assets in terms of both geometry and texture accuracy.

Nonetheless, the model still faces limitations: (1) Even though the $\pi^3$ branch provides geometrically informative features for the 3D branch, in cases where the test view configuration deviates from the training distribution, it may actually disrupt the 3D branch and lead to degraded performance. (2) Due to limited resources, our training utilizes TRELLIS-generated training data, which do not contain lighting or view-dependent visual effects. Directly applying our model to in-the-wild data can lead to degraded performance. (3) The TRELLIS VAE decoders are frozen in our paper, which means the generation quality is limited by the pretrained TRELLIS latent distribution. Please see the supplementary material for a more in-depth discussion and future directions.

\section*{Acknowledgements}
This work is supported by the National Science Foundation of China (NSFC) under Grant Number 62125107.

{
    \small
    \bibliographystyle{ieeenat_fullname}
    \bibliography{my-references}
}


\renewcommand{\thesection}{S\arabic{section}}
\renewcommand{\thefigure}{S\arabic{figure}}
\renewcommand{\thetable}{S\arabic{table}}

\setcounter{section}{0}

\section{Implementation and Evaluation Details}

\subsection{Block Matching Algorithm}

In Sec.~3.3, we designed a block matching strategy which ultimately led to the exact matching Fig.~2. Here, we explain the exact algorithm to derive the specific matching.

According to our matching principle that each global $\pi^3$ block must be matched with a TRELLIS block, we compute a uniform index injection from all the 18 global blocks of $\pi^3$ into the 24 blocks of TRELLIS. Let $P_l (l=0,\cdots,35)$ and $T_j (j=0,\cdots,24)$ denote the blocks of $\pi^3$ and TRELLIS, respectively. Then a simple computation according to the rule above yields the following one-to-one matching: $T_0,T_1,T_2,T_4,T_5,T_6,T_8,T_9,T_{10},T_{12},T_{13},T_{14},T_{16},T_{17}$,\\$T_{18},T_{20},T_{21},T_{23}$ and $P_1,P_3,P_5,P_7,P_9,P_{11},P_{13},P_{15},P_{17}$,\\$P_{19},P_{21},P_{23},P_{25},P_{27},P_{29}$,$P_{31}$,$P_{33},P_{35}$. Now there remain 6 unmatched TRELLIS blocks. The matched pairs become type-C mixtures. Note that there is only a unique way to insert them into the original sequence in an order-preserving way. For example, to insert $T_3$ between matched pairs $(T_2,P_5)$ and $(T_4,P_7)$, we must match $T_3$ and $P_6$. The same goes for all unmatched TRELLIS blocks, giving us type-B mixtures between $T_3,T_7,T_{11},T_{15},T_{19},T_{22}$ and $P_6,P_{12},P_{18},P_{24},P_{30},P_{34}$. Finally, the remaining unmatched $\pi^3$ blocks become type-A blocks. Note that there might be other plausible matchings, but exhausting them is neither practical nor our main focus.

\subsection{Camera Intrinsics Estimation}

In the alignment evaluation, all generated models are rendered back to the input view for evaluation, which requires estimating camera intrinsics. ReconViaGen~\citep{chang2025reconviagen} uses VGGT~\citep{wang2025vggt} which already estimates intrinsics. Our model is based on $\pi^3$~\citep{wang2025pi3} which predicts only extrinsics and local point map. To estimate the intrinsic matrix $K$, we solve for $f_x,f_y,c_x,c_y$ from the following equation:
\begin{eqnarray}
    u&=&f_xx/z+c_x;\\
    v&=&f_yy/z+c_y,
\end{eqnarray}
which is equivalent to
\begin{eqnarray}
    \left[\begin{matrix}
        x/z & 0 & 1 & 0\\
        0 & y/z & 0 & 1\\
    \end{matrix}\right]
    \left[\begin{matrix}
        f_x\\
        f_y\\
        c_x\\
        c_y
    \end{matrix}\right]=
    \left[\begin{matrix}
        u\\
        v
    \end{matrix}\right].
\end{eqnarray}
Here, $(u,v)$ and $(x,y,z)$ ranges over all foreground pixels and 3D points in the output point map of $\pi^3$, leading to an overdetermined equation which we solve in the least-squares sense. This intrinsics is used as an initial value and then refined by differentiable rendering using DRTK~\citep{pidhorskyi2024drtk}. During the differentiable rendering refinement, we parameterize the camera focal as $f=\exp(l)$ and use quaternions to represent camera rotation. The optimization uses the Adam optimizer~\citep{kingma2014adam} and runs for 2000 steps with a learning rate of $10^{-2}$. We apply early stopping if the loss has not decreased for 100 steps.

Note that SAM3D~\citep{sam3dteam2025sam3dobjects} only predicts the object in the camera space without intrinsics. Unlike $\pi^3$, SAM3D does not have a pixel-wise 2D-3D correspondence. Hence, we match the extreme values of $x/z$ and $y/z$ to the input image foreground bounding box to estimate the intrinsics and extrinsics for SAM3D.

\subsection{Dataset}
To train our model, we use the TRELLIS-500k dataset~\citep{xiang2025trellis}, which is composed of selected models from Objaverse-XL~\citep{objaverseXL}, ABO~\citep{collins2022abo} and HSSD~\citep{khanna2023hssd}. Our joint 3D generation and camera estimation pipeline requires ground truth sparse structure latents and structured latents, together with their paired multi-view renderings and camera parameters. However, following the full dataset processing pipeline in TRELLIS requires significant computational resources. Thus, we use only 8-view renderings of each item to directly generate these paired data samples using TRELLIS~\citep{xiang2025trellis}. Finally, after excluding a part of models that either take too long to render or cause generation failures, we generated 404354 objects, each paired with their GT latent codes and 32 renderings densely covering the upper hemisphere with varying focal lengths.

\subsection{Model and Training}
At each training step, we randomly sample an object and 4 views from its 32 renderings. To better cover different view configurations, we use a mixture of the following 3 view sampling modes. (1) Fully random sampling: we randomly choose 4 views without replacement with a uniform distribution. (2) Nearest-view sampling: we first randomly select one view, and then sample its nearest 3 views (``nearest'' in terms of camera positions). (3) Farthest-view sampling: we start with a random view and use farthest point sampling~\citep{qi2017pointnetplusplus} (in terms of camera positions). The probabilities of choosing these three modes are 0.2, 0.4, 0.4, respectively. For training, we use pretrained weights of the TRELLIS sparse structure flow model for our 3D branch, and the pretrained weights of $\pi^3$ for the 2D branch. For the MoT self-attention modules, we load pretrained weights from either TRELLIS or $\pi^3$ whenever possible. During training, we only activate the following parameters: (1) all parameters without a pretrained weight; (2) all parameters of cross-attention modules and MoT self-attention modules. Our core model involved in training (excluding the DINOv2 encoder and the sparse structure VAE decoder) contains 1.71B parameters in total and 839.78M trainable parameters.

We use a learning rate of $10^{-4}$ with cosine scheduling and train for 400k steps with a batch size of 16. The training runs on 16 NVIDIA-A100 GPUs and takes about one week.

\subsection{Evaluation Settings for Baselines}

For ReconViaGen~\citep{chang2025reconviagen}, we use an open-source implementation~\citep{estheryang11reconviagen}. For SAM3D~\citep{sam3dteam2025sam3dobjects}, we use an open-source training-free extension to the multi-view setting~\citep{li2026mvsam3d}, which adopts weighted multi-diffusion to better fuse multi-view information. To adapt TRELLIS and UniLat3D to multi-view inputs, we follow the official repositories to use the \verb|run_multi_image| API, which adopts stochastic image condition sampling to inject multi-view information. Note that the comparison is fair in the sense that (1) both TRELLIS and UniLat3D use the same input views as ours; (2) Our stage-2 model also uses stochastic sampling. The only difference in conditioning is that our stage-1 model uses all-view tokens, which we had attempted for the baselines for fairness, but we found this led to worse baseline performance, so we kept their stochastic sampling.

In our evaluation of geometry and texture accuracy, we need to align all generated shapes to GT ones. However, since generated shapes may not have the exact orientation and scale as the GT ones, we use a heuristic method to achieve this alignment as follows.

For methods that come with camera pose estimations (ours, ReconViaGen and SAM3D), we first compute a rotation between predicted poses and GT poses. This rotation is then applied to the generated object to initialize a good orientation. Then we run ICP from this initialization to get a similarity transformation that provides a more accurate alignment.

For methods that do not estimate pose (TRELLIS~\citep{xiang2025trellis}, UniLat3D~\citep{wu2025unilat3d}), we remark that both of them has $+z$ as the ``up'' direction, and their ``front`` direction is aligned to either the $x$-axis or the $y$-axis. Observing this, we first try out 4 different orientations in the $xy$-plane and find the one with minimal Chamfer distance. Then ICP is run from this initialization to get the final alignment geometry.

\section{Extended Evaluations}

\subsection{Evaluation of Model Components}

In this section we show the effectiveness of several of our technical choices. Note that our stage-1 model is a minimal architecture in the sense that we cannot ablate a part without breaking the whole pipeline. For example, if we remove either the TRELLIS branch, the $\pi^3$ branch or the alignment transformation branch, we wouldn't have aligned voxels and points for stage-2 generation. Also, if we remove the MoT self-attentions, there wouldn't be information exchange to make the aligned training well-defined. Thus, instead of directly ablating the individual branches in our mixed model, we use the following experiments to demonstrate the effectiveness of our mixture design as a whole.

To evaluate the effectiveness of our aligned voxel generation and point map prediction, we compare our stage-1 model and that of TRELLIS, using the "Geometry and texture accuracy" evaluation protocol in Sec.~4.2 to first align generations to GT before computing metrics. The results are reported in Table~\ref{table:supp:stages-eval}. Our stage-1 model produces input-consistent voxels, leading to notable metric improvements even with the TRELLIS stage-2 model. Furthermore, since the stage-1 generation already determines the coarse geometry, our stage-2 generation is more effective for texture alignment. Note that the combination of TRELLIS stage-1 plus our stage-2 is not feasible since TRELLIS does not produce voxel-aligned point maps, so this combination is not evaluated.

\begin{table}
\caption{Separate evaluation of our stage-1 model and our stage-2 model.\label{table:supp:stages-eval}}
\centering
\footnotesize
\begin{tabular}{cc|cccc}
\toprule
 & & \multicolumn{4}{c}{Toys4k} \\
\midrule
Stage-1 & Stage-2 & PSNR & SSIM & LPIPS & CD ($\times 10^{-3}$)  \\
\midrule
TRELLIS & TRELLIS & 23.930 & 0.9311 & 0.055 & 7.9986\\
Ours & TRELLIS    & 26.975 & 0.9527 & 0.035 & \textbf{0.7259}\\
Ours & Ours       & \textbf{27.177} & \textbf{0.9551} & \textbf{0.033} & 0.7419 \\
\midrule
& & \multicolumn{4}{c}{GSO} \\
\midrule
Stage-1 & Stage-2 & PSNR & SSIM & LPIPS & CD ($\times 10^{-3}$) \\
\midrule
TRELLIS & TRELLIS & 22.420 & 0.9034 & 0.079 & 4.5658 \\
Ours & TRELLIS    & 24.399 & 0.9199 & 0.062 & 0.8010 \\
Ours & Ours       & \textbf{24.826} & \textbf{0.9271} & \textbf{0.055} & \textbf{0.7945} \\
\bottomrule
\end{tabular}
\end{table}

To show that our MoT-based architecture also conversely benefits the point map prediction, we evaluate the point map accuracy of our model and that of $\pi^3$, using pixel-wise error (PE) and Chamfer distance (CD), both computed after a similarity transformation alignment. The results are shown in Table~\ref{table:supp:pointerror}.

\begin{table}
\caption{Evaluation pf point map errors ($\times10^{-3}$).\label{table:supp:pointerror}}
\centering
\footnotesize
\begin{tabular}{c|cc|cc}
\toprule
 & \multicolumn{2}{c|}{Toys4k} & \multicolumn{2}{c}{GSO} \\
\midrule
Method & PE  & CD  & PE  & CD  \\
\midrule
$\pi^3$ & 9.441 & 3.753 & 6.723 & 2.896 \\
Ours    & \textbf{2.285} & \textbf{0.527} & \textbf{3.729} & \textbf{1.258} \\
\bottomrule
\end{tabular}
\end{table}

\begin{figure*}
  \centering
   \includegraphics[width=0.8\linewidth]{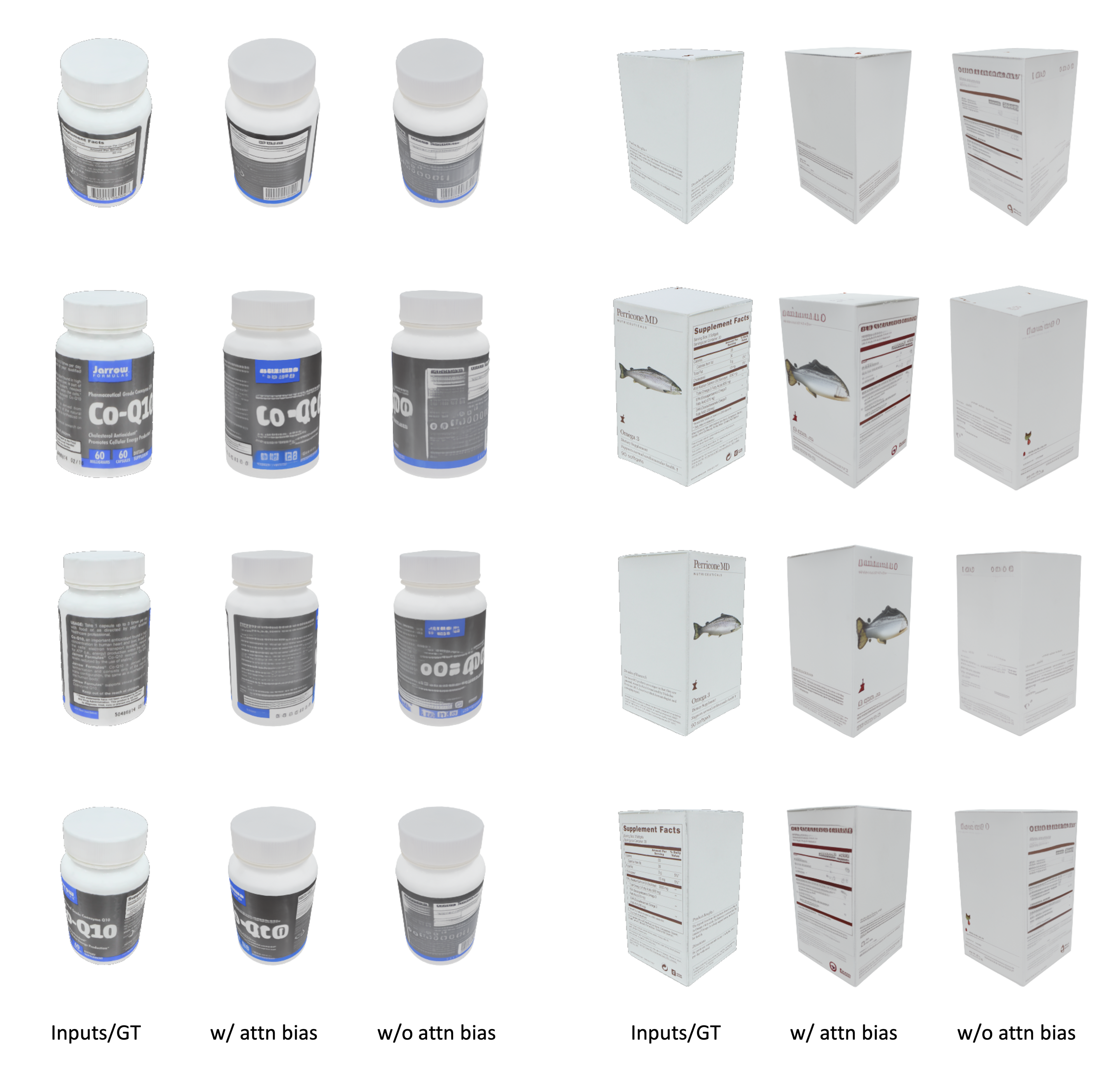}
   \caption{Qualitative studies of ablating our proposed attention bias.}
   \label{fig:supp:abl}
\end{figure*}

Finally, we evaluate the usage of attention biases and camera refinement. Table~\ref{table:supp:abl} shows the metrics of our method after removing camera refinement and attention bias. Even though the metric improvement brought by using our attention bias seems minor, the improvement can be more clearly observed with qualitative examples. Fig.~\ref{fig:supp:abl} shows cases where our attention bias notably improves texture alignment. Note that this type of improvement often happens to rotationally symmetric objects with asymmetric texture. However, in both Toys4k~\citep{stojanov21cvpr} and GSO~\citep{downs2022gso} there are only limited cases like this. This is why our attention bias has a marginal metric improvement but remains important and effective in improving texture alignment in multi-view settings. To further verify this, we manually annotated items in the GSO dataset as "Asymmetric-Geometry (AG)", "Symmetric-Geometry Asymmetric-Texture (SGAT)" and "Symmetric-Geometry Symmetric-Texture (SGST)". Note that we only consider rotational symmetry (balls, square boxes, bottles, etc.), since mirror symmetry can be visually disambiguated from only the geometry. Table~\ref{table:supp:symasym} shows the metric results. Our proposed attention bias in stage-2 is indeed most effective on SGAT objects.

The ablations on attention bias and camera refinement are shown in Fig.~\ref{table:supp:abl}.

\begin{table}
\caption{Ablations of attention bias and camera refinement.\label{table:supp:abl}}
\centering
\footnotesize
\begin{tabular}{cc|ccc}
\toprule
Attn Bias & Refine      & PSNR & SSIM & LPIPS \\
\midrule
\multicolumn{2}{c|}{Dataset} & \multicolumn{3}{c}{Toys4k} \\
\midrule
No  & No                & 25.014 & 0.9397 & 0.042 \\
Yes & No                & 25.163 & 0.9422 & 0.039 \\
Yes & Yes               & 28.144 & 0.9621 & 0.030 \\
\midrule
\multicolumn{2}{c|}{Dataset} & \multicolumn{3}{c}{GSO} \\
\midrule
No  & No                & 23.020 & 0.9039 & 0.078 \\
Yes & No                & 23.350 & 0.9107 & 0.070 \\
Yes & Yes               & 25.031 & 0.9281 & 0.057 \\
\bottomrule
\end{tabular}
\end{table}

\begin{table}
\caption{Ablations of attention bias and camera refinement on symmetric and asymmetric objects in the GSO dataset.\label{table:supp:symasym}}
\centering
\footnotesize
\begin{tabular}{cc|ccc}
\toprule
 & & \multicolumn{3}{c}{AG (645 objects)} \\
\midrule
Attn Bias & Refine      & PSNR & SSIM & LPIPS \\
\midrule
No  & No                & 24.341 & 0.9293 & 0.052 \\
Yes & No                & 24.515 & 0.9336 & 0.049 \\
Yes & Yes               & \textbf{25.675} & \textbf{0.9434} & \textbf{0.044} \\
\midrule
 & & \multicolumn{3}{c}{SGAT (279 objects)} \\
\midrule
Attn Bias & Refine      & PSNR & SSIM & LPIPS \\
\midrule
No  & No                & 20.849 & 0.8578 & 0.112 \\
Yes & No                & 21.654 & 0.8724 & 0.093 \\
Yes & Yes               & \textbf{23.207} & \textbf{0.8958} & \textbf{0.077} \\
\midrule
 & & \multicolumn{3}{c}{SGST (97 objects)} \\
\midrule
Attn Bias & Refine      & PSNR & SSIM & LPIPS \\
\midrule
No  & No                & 20.303 & 0.8644 & 0.150 \\
Yes & No                & 20.328 & 0.8661 &0.149 \\
Yes & Yes               & \textbf{25.923} & \textbf{0.9178} & \textbf{0.094} \\
\bottomrule
\end{tabular}
\end{table}

\subsection{Runtime Report}

For all our experiments, we use an input image resolution of 518, and we run 50 steps for stage-1 generation, and 25 steps for stage-2 generation. With 4 input views, on a single NVIDIA-A100 GPU, our stage-1 generation takes around 30s, and our stage-2 generation generally takes around 3$\sim$10s, depending on the number of voxels generated in stage 1. The attention bias computation generally takes $<$10s. The camera refinement process takes no more than 10s per view. Note that this is not needed if accurate poses are not required.

\section{Extended Discussions on Limitations}

In this section, we provide further discussions on the limitations of our method and possible future directions.

\paragraph{Degradation caused by view configuration} While our method utilizes $\pi^3$ to provide geometrically informative features, different view configurations can impact how well the $\pi^3$ branch aligns different view points. Generally, the best performance can be obtained if all views are directly looking at the center of the object with a zero roll angle. However, in real-world scenarios it is difficult to strictly follow this rule, and therefore performance degradation may happen if the input view configuration deviates too much from its training distribution. Future work should attempt training with a more diverse view distribution or utilize data augmentation to improve the robustness.

\begin{figure}[h]
  \centering
   \includegraphics[width=\linewidth]{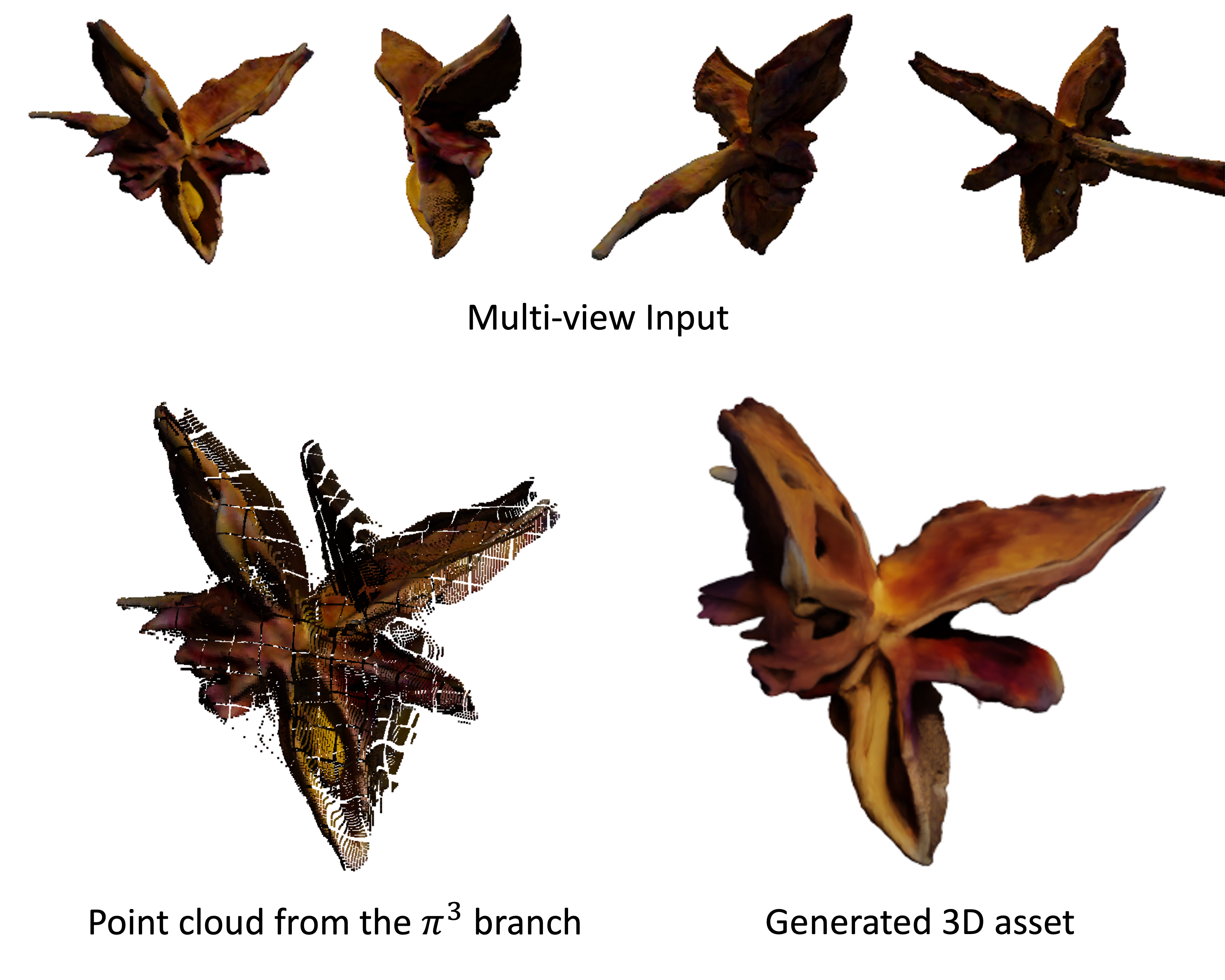}
   \caption{A failure case caused by non-uniform lighting.}
   \label{fig:supp:failure2}
\end{figure}

\paragraph{Limitations of using generated data} As mentioned in Sec.~4.1, our model is trained on TRELLIS-generated data due to the extreme high cost of running the original data processing pipeline of TRELLIS. These generated models do not contain lighting or view-dependent visual effects. As a result, our method has degraded performance for datasets with highly non-uniform lighting, e.g., OmniObject3D~\citep{wu2023omniobject3d}, or specular objects. An example is shown in Fig.~\ref{fig:supp:failure2}, where dark shadows appear on the side views of the object. Our $\pi^3$ branch, not trained on these lighting conditions, cannot correctly match the image features through its attention mechanism. As a result, the final generation can neither benefit much from features of the $\pi^3$ branch, nor from the attention bias tuning which requires accurate overlap. We believe training with high quality data containing diverse lighting can remedy this issue.

\paragraph{Frozen TRELLIS decoders} Our model keeps all the TRELLIS VAE decoders frozen. In other words, the latent spaces of TRELLIS impose an upper bound for our generation quality. For certain types of textures, e.g., texts or logos on commercial products, even reconstructing them using the TRELLIS VAE is problematic. This also limits the faithfulness of geometry or texture preservation of our model. However, since MoT architectures can be easily inserted into different transformers, a possible future direction is to extend our designs to more powerful backbones~\citep{hunyuan3d22025tencent,lai2025hunyuan3d25highfidelity3d,xiang2025trellis2}.

\end{document}